%% file: main.tex
\documentclass[10pt,twocolumn,letterpaper]{article}

\usepackage[pagenumbers]{cvpr} % To force page numbers, e.g. for an arXiv version

\usepackage{pifont}
\usepackage{txfonts}
\usepackage{multirow}
\usepackage{amssymb}

\definecolor{cvprblue}{rgb}{0.21,0.49,0.74}
\usepackage[pagebackref,breaklinks,colorlinks,allcolors=cvprblue]{hyperref}

\def\paperID{1002} % *** Enter the Paper ID here
\def\confName{CVPR}
\def\confYear{2026}

\title{Highly Efficient Test-Time Scaling for T2I Diffusion Models with Text Embedding Perturbation}

\author{Hang Xu$^{1}$  \quad Linjiang Huang$^{2}$ \quad  \quad Feng Zhao$^{1}$\thanks{Corresponding author}\\
$^{1}$MoE Key Lab of BIPC, USTC $^{2}$Beihang University\\
}

\begin{document}
\maketitle
\input{text/0_abstract}    
\input{text/1_intro}

\input{text/2_related}

\input{text/3_motivation}

\input{text/4_method}

\input{text/5_exp}

\input{text/6_conclusion}

{
    \small
    \bibliographystyle{ieeenat_fullname}
    \bibliography{main}
}


\end{document}

%% file: text/0_abstract.tex
\begin{abstract}
Test-time scaling (TTS) aims to achieve better results by increasing random sampling and evaluating samples based on rules and metrics. However, in text-to-image(T2I) diffusion models, most related works focus on search strategies and reward models, yet the impact of the stochastic characteristic of noise in T2I diffusion models on the method's performance remains unexplored.  
In this work, we analyze the effects of randomness in T2I diffusion models and explore a new format of randomness for TTS: text embedding perturbation, which couples with existing randomness like SDE-injected noise to enhance generative diversity and quality. We start with a frequency-domain analysis of these formats of randomness and their impact on generation, and find that these two randomness exhibit complementary behavior in the frequency domain:  spatial noise favors low-frequency components (early steps), while text embedding perturbation enhances high-frequency details (later steps), thereby compensating for the potential limitations of spatial noise randomness in high-frequency manipulation. Concurrently, text embedding demonstrates varying levels of tolerance to perturbation across different dimensions of the generation process. Specifically, our method consists of two key designs: (1) Introducing step-based text embedding perturbation, combining frequency-guided noise schedules with spatial noise perturbation. (2) Adapting the perturbation intensity selectively based on their frequency-specific contributions to generation and tolerance to perturbation. Our approach can be seamlessly integrated into existing TTS methods and demonstrates significant improvements on multiple benchmarks with almost no additional computation. Code is available at \href{https://github.com/xuhang07/TEP-Diffusion}{https://github.com/xuhang07/TEP-Diffusion}.

\end{abstract}

%% file: text/1_intro.tex
\section{Introduction}
\label{sec:intro}
\begin{figure}[t]
	\centering
	\includegraphics[width=1\linewidth]{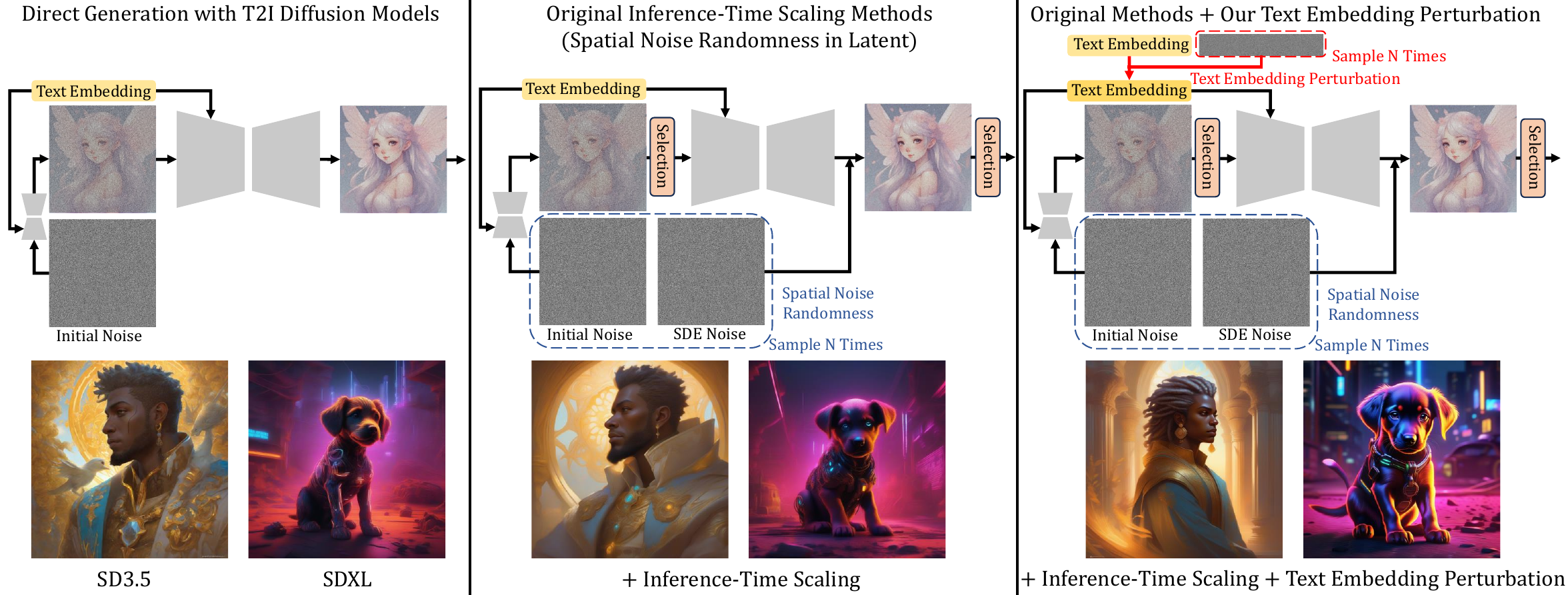}
	\caption{Top: Comparison of text embedding perturbation with previous randomness. Bottom: The corresponding generated images of the Top. Our method is plug-and-play.
 % It can be integrated into current methods at fewer more costs.
 } 
	\label{fig:first}
\end{figure}

Diffusion models start from random noise and have demonstrated impressive generative capabilities in text-to-image(T2I) generation. However, due to the inconsistency in their training-inference paradigm, where multiple noise-to-data mappings are learned during training but only a single noise is used during inference, the full potential of diffusion models in generation remains untapped. Therefore, inspired by the test-time scaling (TTS) techniques in LLMs~\cite{muennighoff2025s1,liu2025inference}, many researchers aim to enhance the generation quality of diffusion models by scaling inference computations during inference~\cite{ma2025inference,singhal2025general}. Specifically, these TTS methods rely on the sampling randomness of diffusion models (like initial noise) to generate multiple candidate samples, evaluate them using reward models, and then employ search strategies to select and further refine the candidates. Therefore, the core components of TTS methods consist of randomness, search strategies, and reward models. 

Research on search strategies and reward models has dominated TTS methods for T2I diffusion models, while randomness and its impact on these methods remain unexplored. Notably, randomness directly affects the size of the search space in TTS methods~\cite{zhang2025inference}. However, most existing works rely solely on spatial random noise introduced in latent space (i.e., SDE), which may not provide a sufficiently large search space. A constrained search space means repeated sampling tends to converge on similar and redundant candidates, leading to ineffective use of computational resources~\cite{puri2025probabilistic}. Therefore, it is meaningful to explore a new format of randomness that can both enhance generative diversity and complement existing spatial noise randomness.

In this paper, we explore a new format of randomness, text embedding perturbation, for TTS methods in T2I diffusion models. While recent studies have utilized text embedding perturbation to generate more diverse images~\cite{sadat2023cads}, they struggle to maintain visual quality and text faithfulness, making them unsuitable for TTS methods (see Fig. \ref{fig:existing_methods}). 
Our experimental analysis attributes this limitation to two key factors: \textbf{(1)} poor complementarity between text embedding perturbation and existing spatial noise randomness, and \textbf{(2)} excessive uniformity in the responses of diffusion model components.
\textbf{First}, from a frequency-domain perspective, we reveal a complementary relationship between text embedding perturbation and spatial noise randomness: while spatial noise randomness primarily affects low-frequency components, text embedding perturbation enhances high-frequency details (Fig. \ref{fig:sde}). This complementarity extends to their joint impact on image quality throughout denoising (Figs. \ref{fig:sde2ode} and Fig. \ref{fig:tep2ntep}). However, prior work may not focus on this synergy, degrading both visual quality and text alignment.
\textbf{Second}, as shown in Fig. \ref{fig:only_tep}, text embeddings demonstrate distinct requirements for and tolerance of perturbation based on the time step, the specific components of the embedding, and the depth of the Cross-Attention layer. Consequently, applying a discriminative perturbation strategy tailored to the needs of these unique dimensions is essential, which is a crucial aspect but has not been explored in existing research related to TTS.
% Our experimental analysis attributes this to poor complementarity with existing spatial noise randomness and excessive uniformity across components.
% Specifically, first, from a frequency-domain perspective, we uncover the complementary relationship between text embedding perturbation and classical spatial noise randomness: spatial noise randomness primarily influences low-frequency components, while text embedding perturbation enhances high-frequency details, also shown in Fig. \ref{fig:sde}. Meanwhile, their impact on image quality exhibits complementarity throughout the entire denoising process (Fig. \ref{fig:sde2ode} and \ref{fig:tep2ntep}). Yet, existing studies have not explored this complementarity, compromising both visual quality and text alignment.
% Second, the CFG (Classifier-Free Guidance)~\cite{ho2022classifier} in T2I diffusion models exhibits varying responses to perturbations across its components. As shown in Fig. \ref{fig:uncond_cond}, CFG shifts generation from unconditional text embeddings toward conditional embeddings, where the former exhibits significantly higher noise tolerance than the latter. However, prior work has not focused on this asymmetry between the two branches, limiting the full potential of text embedding perturbation.

Therefore, to fully unlock the potential of text embedding perturbation as a new format of randomness, we propose the Text Embedding Perturbation (TEP) framework (see Fig. \ref{fig:framework}), which integrates text embedding perturbation into existing TTS methods. Our framework features two key designs: \textbf{(1)} We introduce step-based text embedding perturbation, combining frequency-guided noise schedules with spatial noise perturbation to further enhance the complementarity between text embedding perturbation and spatial noise randomness. \textbf{(2)} A mild perturbation is applied to the conditional text embedding and deeper layers, while a stronger perturbation is applied to the unconditional text embedding and shallower layers, better aligning with their distinct response to perturbation. Moreover, to preserve core textual semantics while inducing randomness, we employ token-wise adaptive perturbation, accounting for the differential importance of semantic components during generation.
Notably, our TEP framework offers a ``\textbf{free lunch}": it seamlessly integrates with all existing TTS methods with negligible additional computational cost (Tab. \ref{tab:ablation}), while significantly boosting their performance ceiling (Fig. \ref{fig:distribution}).

We summarize the contributions as follows: \textbf{(1)} We introduce a novel format of randomness, text embedding perturbation, for TTS, and systematically analyze its integration properties with spatial noise randomness within TTS methods. \textbf{(2)} We propose the TEP framework that incorporates temporal perturbation strength scheduling based on a frequency-based mechanism, and spatially distinctive perturbation through branch-wise response modulation. \textbf{(3)} We show the effectiveness and plug-and-play of our framework across existing TTS methods with negligible overhead.

%% file: text/2_related.tex
\section{Related Works}
\label{sec:related}
\paragraph{Reward Alignment in Diffusion Models} optimizes generation to maximize reward model evaluations. Current approaches fall into two categories: fine-tuning-based methods~\cite{wallace2024diffusion,liu2025flow} that adapt models using preference data or reward gradients (computationally costly and inflexible), and more flexible TTS techniques that require no retraining. The latter can be further divided into gradient-based~\cite{song2020score,bansal2023universal} and sampling-based methods~\cite{ma2025inference,singhal2025general}, with sampling approaches offering particular advantages: they don't require differentiable reward models, making them flexible to use.
\begin{figure}[t]
    \centering
    \subfloat[Comparison of randomness.\label{fig:diversity_cfg}]{
        \includegraphics[width=0.45\linewidth]{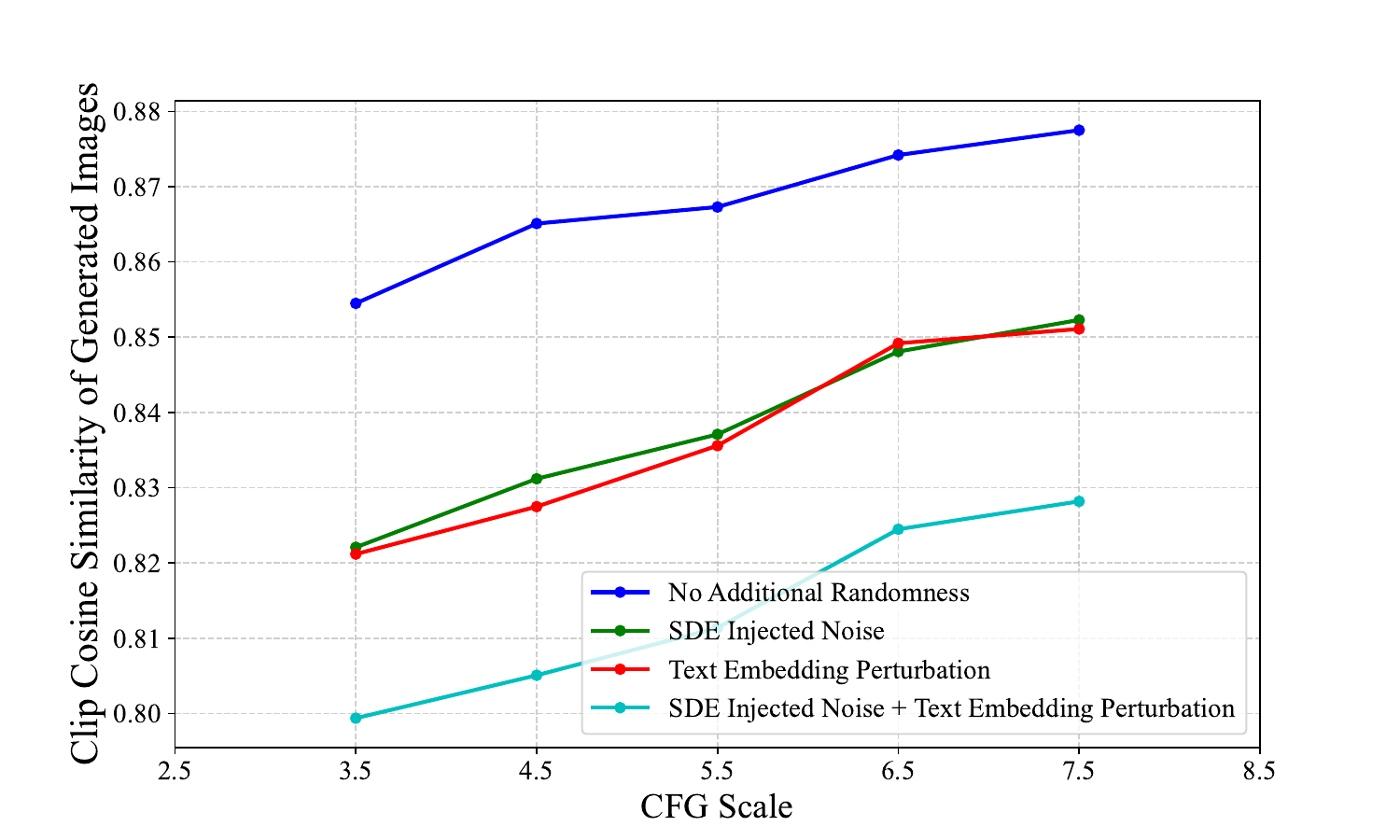}
    }
    \hfill
    \subfloat[Image quality w/ and w/o CADS.\label{fig:cads}]{
        \includegraphics[width=0.45\linewidth]{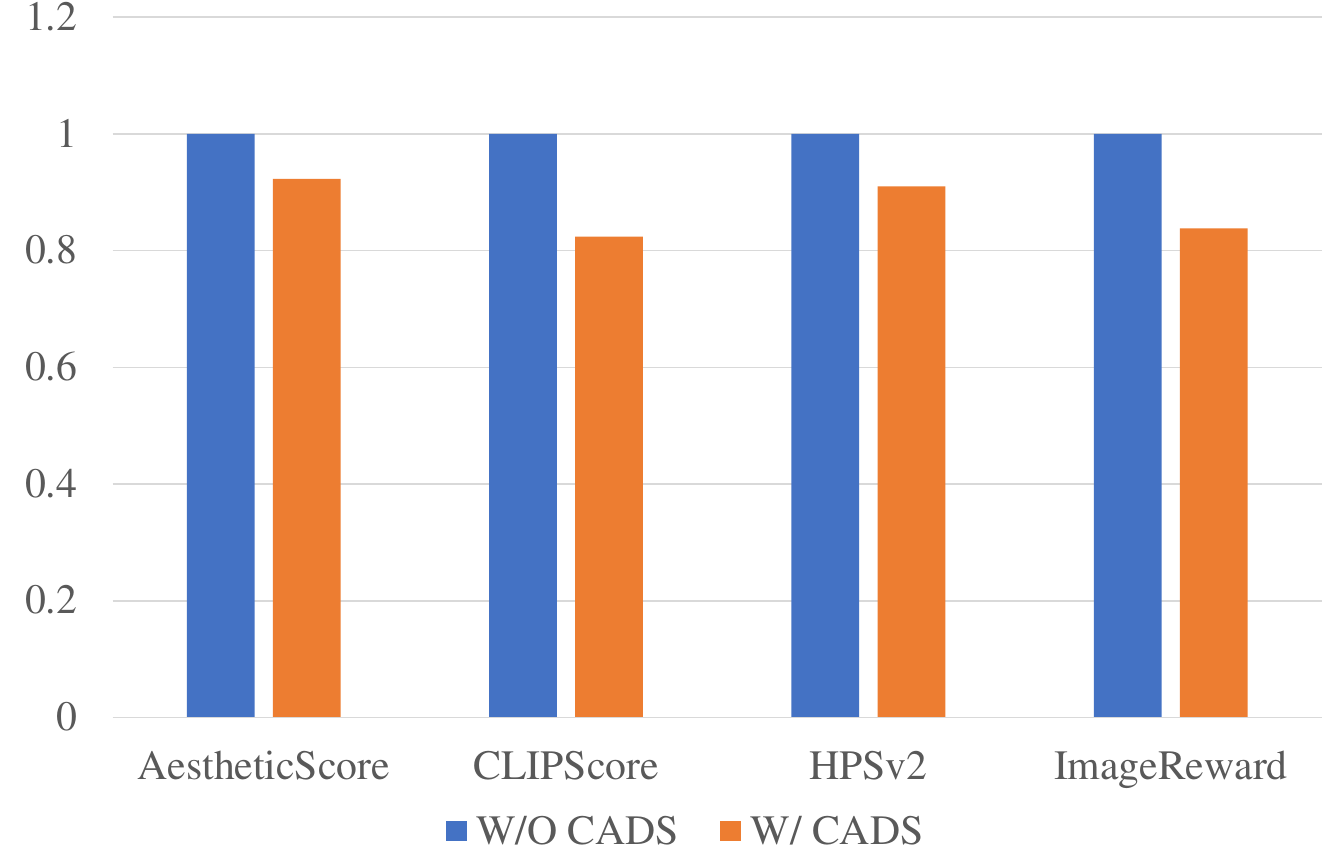}
    }
    
    \caption{In (a), we demonstrate that text embedding perturbation consistently enhances generation diversity across all CFG scales, highlighting its potential as a novel format of randomness. However, in (b), we reveal that existing approaches(CADS) incorporating text embedding perturbation for better generative diversity are incompatible with TTS in T2I diffusion models, as they may not maintain image quality. \textbf{We use SD3.5 with ImageReward for evaluations, and results of more backbones are in Appendix.}}
    \label{fig:existing_methods}
    \vspace{-10pt}
\end{figure}
\vspace{-5pt}
\paragraph{Sampling-Based TTS Methods} operate through two core mechanisms: the diffusion model's inherent stochastic sampling and reward-guided filtering. These approaches can be categorized by their randomness sources into three types: (1) ODE-based methods~\cite{ma2025inference} that rely solely on initial noise, directly denoising it for final selection; (2) Particle sampling~\cite{singhal2025general,singh2025code} that injects additional SDE noise during denoising to explore diverse trajectories, employing either best-of-N or importance sampling strategies for high-reward potential selection; and (3) Resampling-based methods~\cite{ma2025inference} that combine ODE processes with additional resampling operations to regenerate high-reward potential. 
\vspace{-5pt}
\paragraph{Text Embedding} serves as a bridge between text and images in T2I diffusion models and has become an indispensable component of modern models~\cite{yu2024uncovering}. Some studies have demonstrated that perturbing text embeddings can enhance the diversity of image generation. For example, CADS~\cite{sadat2023cads} adds Gaussian noise to the condition and gradually anneals it during generation, resulting in improved diversity.

%% file: text/3_motivation.tex
\section{Motivation}
\label{sec:motivation}

\begin{figure}[t]
    \centering
    \subfloat[The impact of switching from SDE to ODE at a specific step.\label{fig:sde2ode}]{\includegraphics[width=0.47\linewidth]{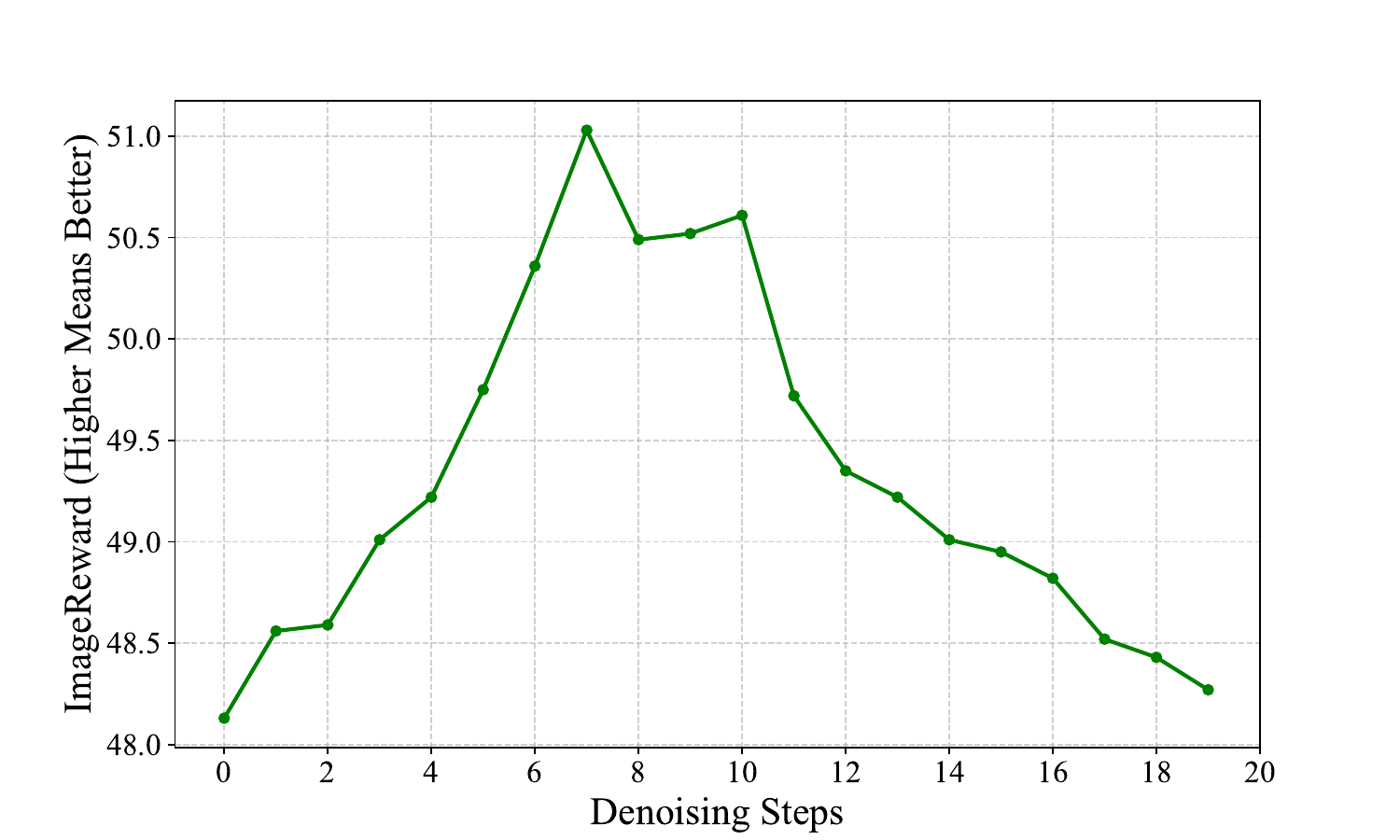}}
    \hfill
    \subfloat[The influence of the high-frequency and the low-frequency components in SDE-injected noise on image quality. \label{fig:frequency}]{
        \includegraphics[width=0.47\linewidth]{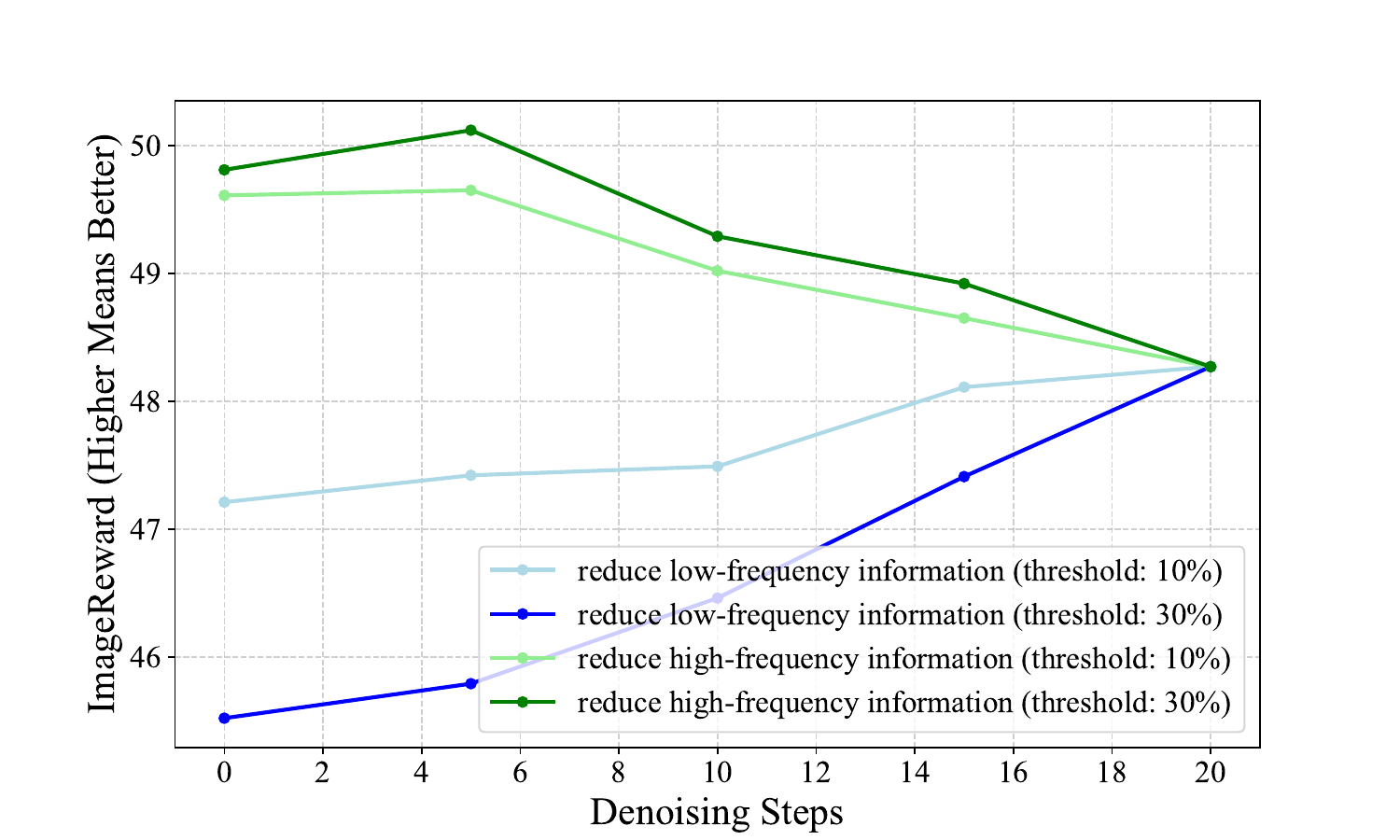}
    }
    \caption{In (a), we switch from SDE to ODE at specific steps. We observe that the noise injected by the SDE in the early steps helps select better-quality images, while in the later steps, it has a negative impact. In (b), we attenuate the SDE-injected noise in the frequency domain at specific steps to analyze the influence of its high- and low-frequency components on generation. We find that the low-frequency one play a crucial role throughout the entire process, whereas suppressing the high-frequency components improves image quality in TTS. \textbf{We use SD3.5 with ImageReward for evaluations, and more backbones' results are in Appendix.} }
\end{figure}

Unless otherwise specified, we employ \textbf{SD3.5} with \textbf{ImageReward} as the evaluation metric. The corresponding analysis results for other mainstream diffusion models, which exhibit similar results, are provided in \textbf{Appendix}.
\subsection{Preliminaries: Generative Diversity and Quality in Diffusion Models}
\label{sec:preliminary}

T2I diffusion models rely on CFG to generate high-quality outputs, which blend the predictions of a conditional model and an unconditional model with a CFG scale $w$:
\begin{equation}
\label{eq:cfg}
    \hat{\epsilon}_t = \epsilon_\theta(x_t, t, E(y_\emptyset)) + w\left( \epsilon_\theta(x_t, t, E(y_c)) - \epsilon_\theta(x_t, t, E(y_\emptyset)) \right)
\end{equation}
where $y_\emptyset$ and $y_c$ represent the null text prompt and the text prompt, respectively. $E$ represents the text encoder.

\begin{figure}[t]
    \centering
    \subfloat[Comparison of influence on images.\label{fig:influence}]{\includegraphics[width=0.45\linewidth]{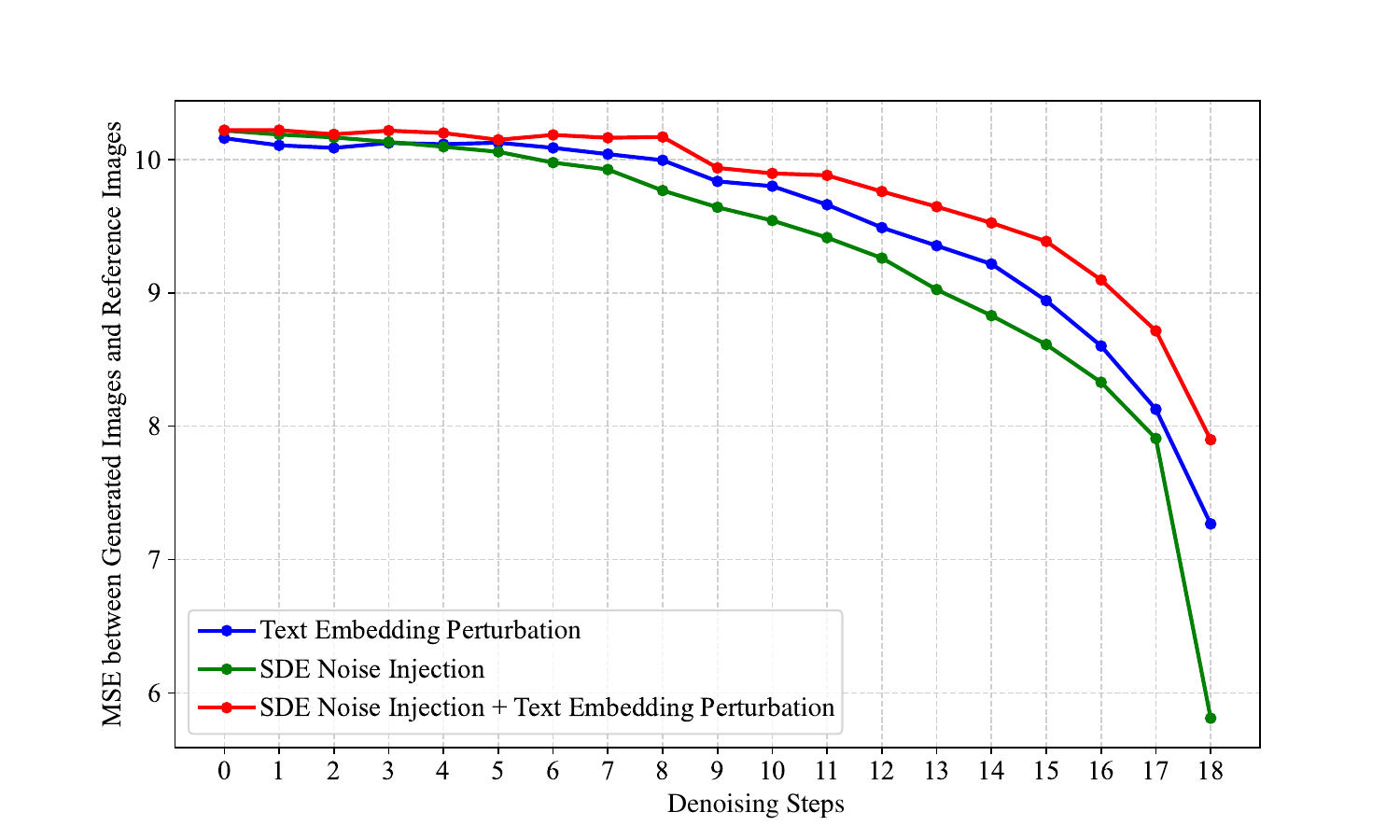}}
    \hfill
    \subfloat[Generative diversity with different randomness.\label{fig:diversity_visual}]{
        \includegraphics[width=0.52\linewidth]{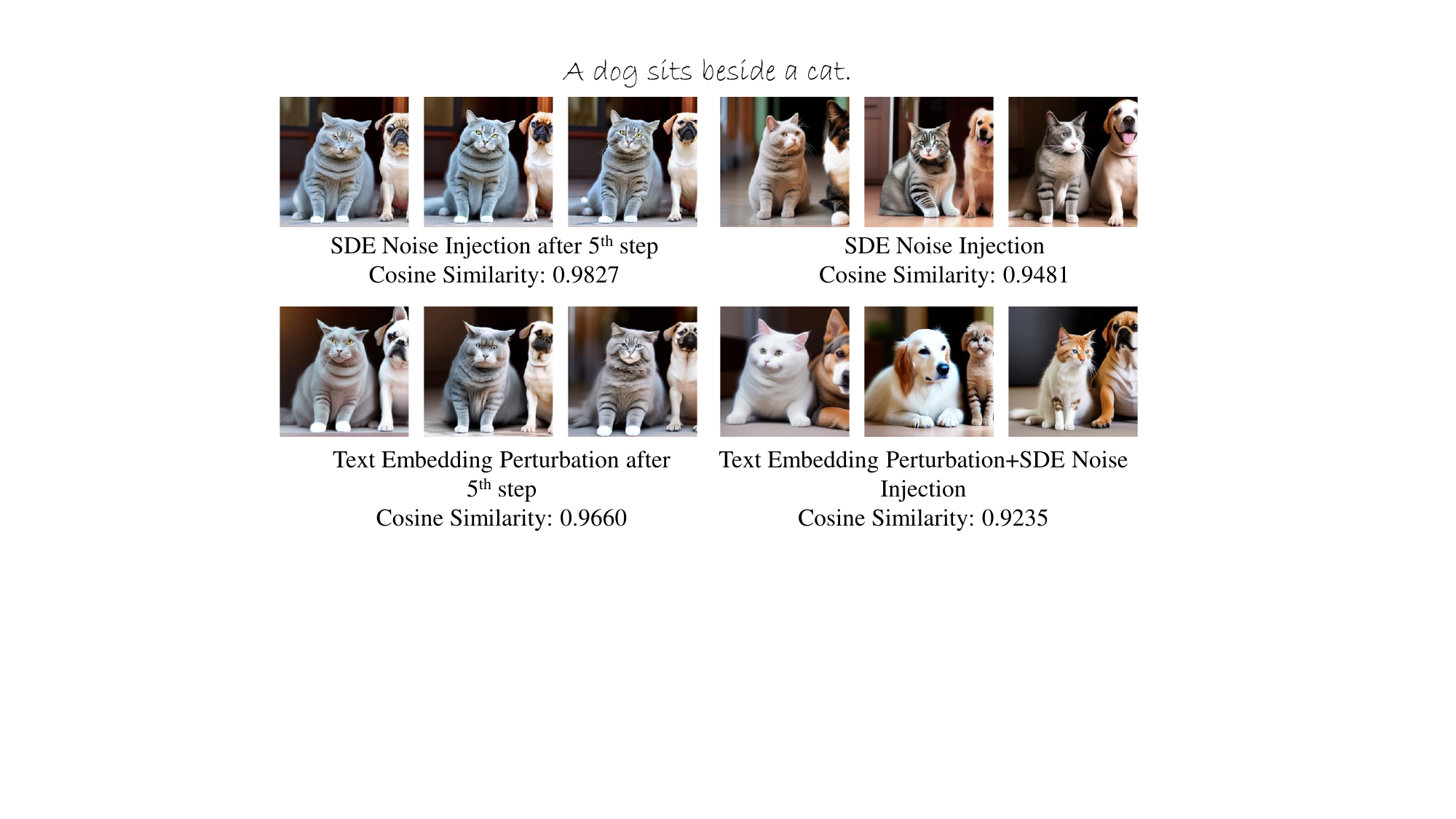}
    }
    
    \caption{\textbf{We demonstrate the differences and complementarity between text embedding perturbation and spatial noise randomness in TTS.} In (a), we only introduced randomness at fixed timesteps while keeping other steps unchanged, and measured the MSE between images generated with and without randomness. Spatial noise randomness has a greater impact during the low-frequency generation phase (early steps), while text embedding perturbation plays a more significant role in the high-frequency phase (late steps). Furthermore, when both randomness formats are combined, the variation in images shows a more pronounced improvement. In (b), we further provide visual results to support the previous discussion. \textbf{We use SD3.5 with ImageReward for evaluations, and results of more backbones are in Appendix.}}
    \label{fig:sde}
\end{figure}

Numerous studies have demonstrated that CFG can impair the generative diversity of models~\cite{moufad2025conditional,koulischer2025feedback}, as we also illustrate in Fig. \ref{fig:diversity_cfg}. Here, we follow ~\citet{cideron2024diversity}, and employ cosine similarity between embeddings of generated samples with the same conditions to quantify the model's generative diversity:
\begin{equation}
\label{eq:sim}
D(\theta) = \mathbb{E}_{\mathbf{y} \sim Y} \left[ \mathbb{E}_{x_1, x_2 \sim p_\theta(\cdot|\mathbf{y})} \left[ s_D(x_1, x_2) \right] \right]
\end{equation}
\begin{equation}
\label{eq:cos}
s_D(x_1, x_2) = 1 - \frac{|E(x_1) \cdot E(x_2)|}{|E(x_1)| |E(x_2)|}
\end{equation}
Where $p_\theta(\cdot|\mathbf{y})$ represents the probability distribution of a certain variable given a text prompt and $E$ represents the embedding model. Here we use CLIP as the embedding model, and we calculate the average diversity value of 1k prompts on SD3.5~\cite{esser2024scaling}. As shown in Fig. \ref{fig:diversity_cfg}, introducing extra randomness to the generation benefits the generation diversity like SDE-injected noise. Therefore, some work begins to introduce other forms of randomness.

Recent studies have demonstrated that perturbing text embeddings can enhance the diversity of generation in diffusion models, which involves adding substantial perturbations to the entire text embedding initially and gradually annealing them during the denoising process~\cite{sadat2023cads}. Since the composition of generated images is primarily shaped by both the text embedding and initial noise in the early steps~\cite{yi2024towards}, applying stronger perturbations to the text embedding during these early stages can enrich image composition, thereby improving generative diversity as shown in Fig. \ref{fig:diversity_cfg}. However, such perturbations severely disrupt semantic information, harming text-image alignment and, to some extent, degrading compositional quality as illustrated in Fig. \ref{fig:cads}. This trade-off is unacceptable for TTS methods. Therefore, we aim to identify an appropriate way to introduce text embedding perturbation into TTS methods by specifically analyzing its impact on generation quality and complementarity with existing randomness (i.e., SDE).

\begin{figure}[t]
    \centering
    \subfloat[Diagram of text embedding perturbation intensity across three dimensions in (b), (c) and (d).\label{fig:differentiation}]{\includegraphics[width=0.50\linewidth]{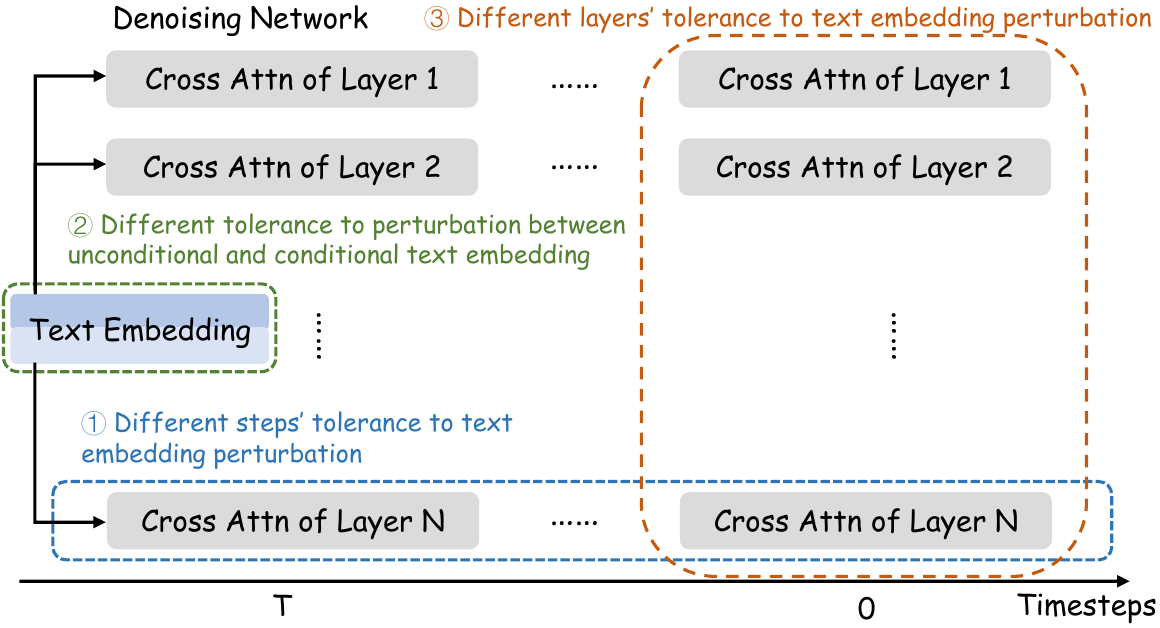}}
    \hfill
    \subfloat[The impact of switching from applying text embedding perturbation to not at a specific step.\label{fig:tep2ntep}]{\includegraphics[width=0.47\linewidth]{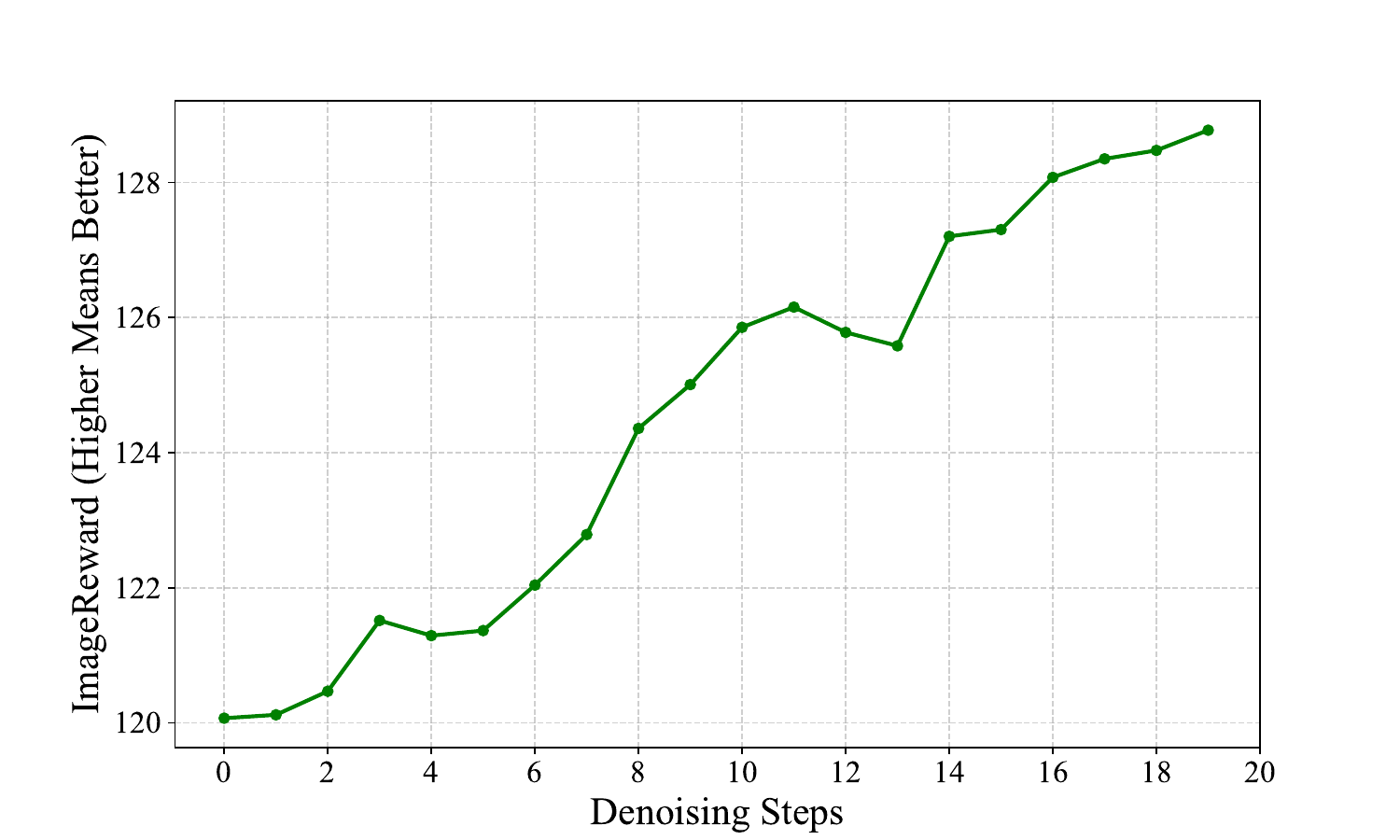}}
    \\
    \subfloat[The tolerance to perturbation of unconditional and conditional text embedding. \label{fig:uncond_cond}]{
        \includegraphics[width=0.46\linewidth]{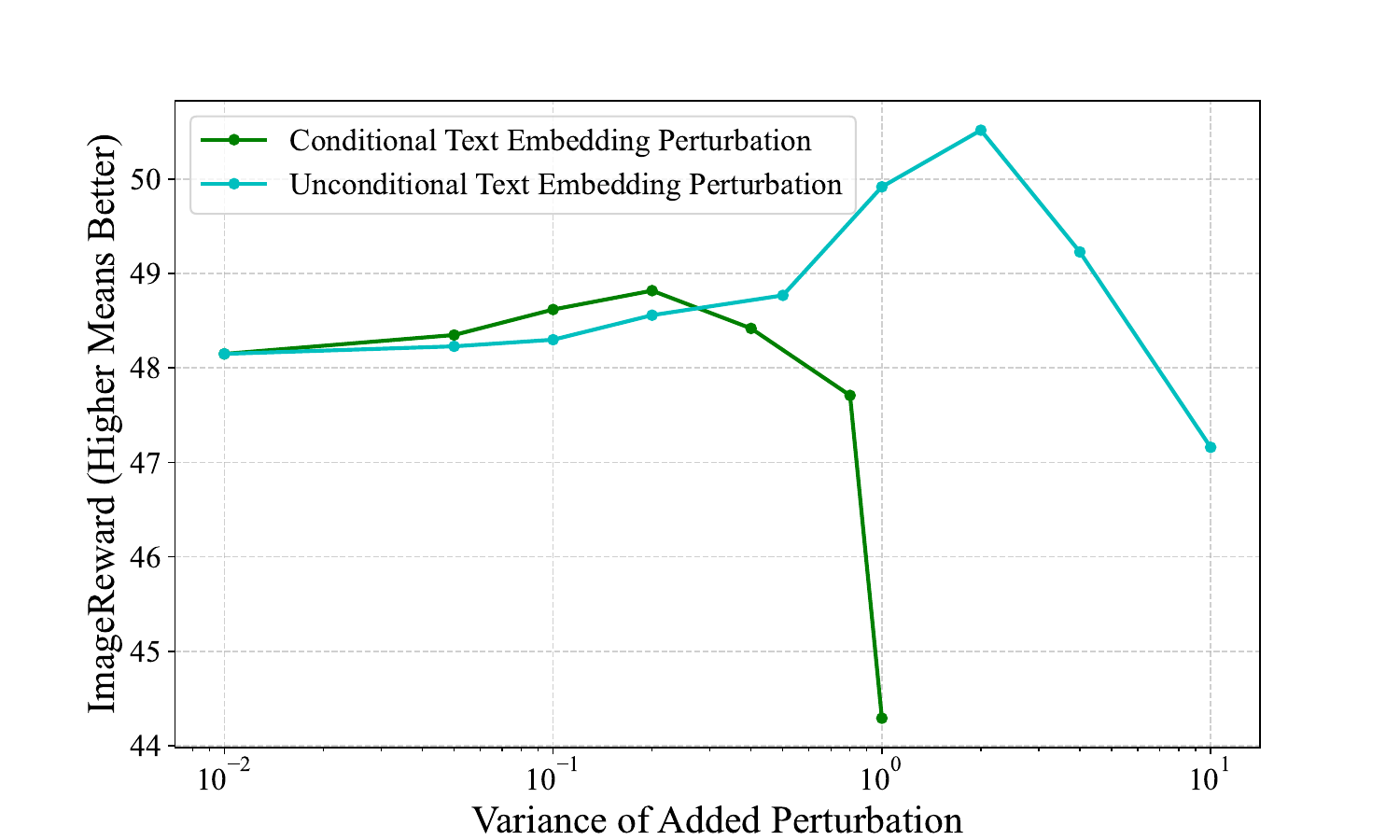}
    }
    \hfill
    \subfloat[Cross attention of different layers' tolerance to text embedding perturbation. \label{fig:layer_perturb}]{
        \includegraphics[width=0.48\linewidth]{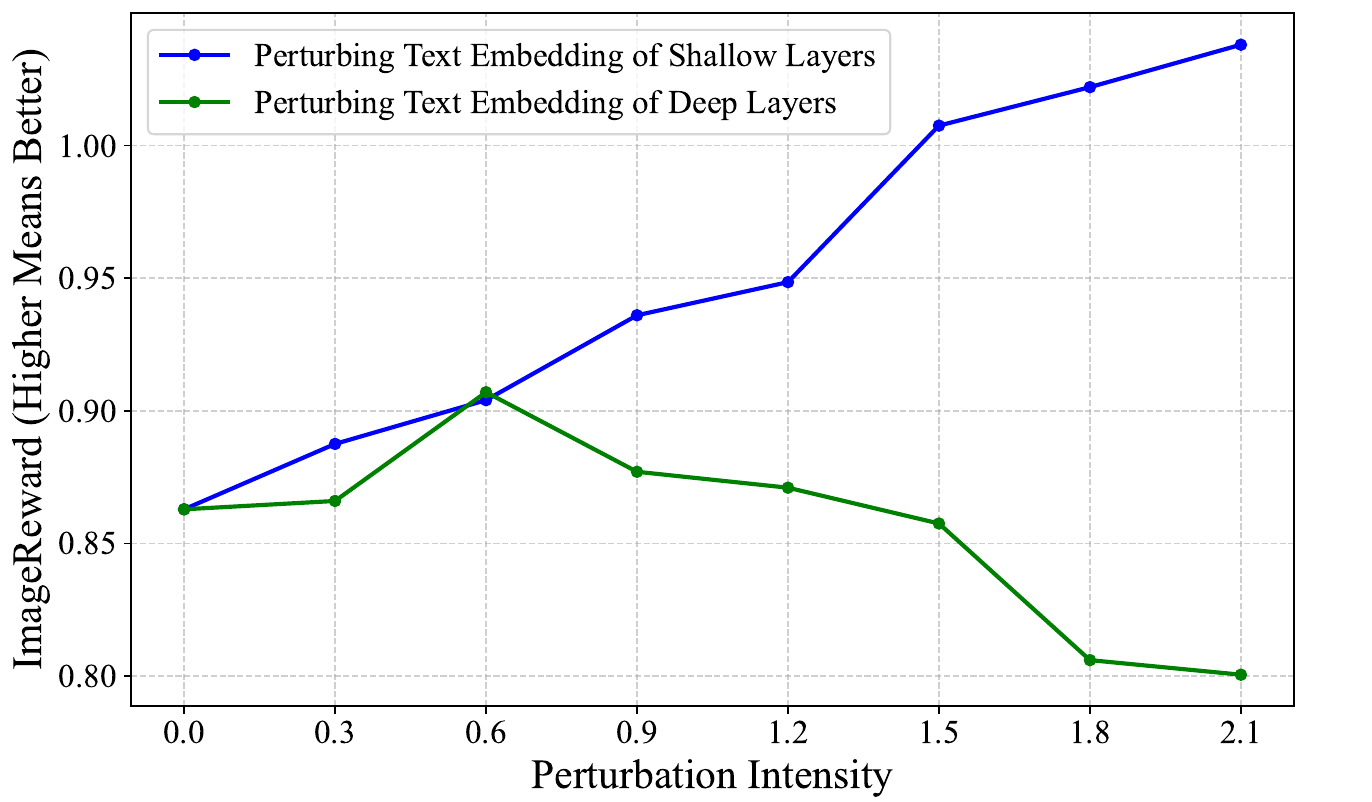}
    }
    
    \caption{(a) shows text embedding perturbation should be applied discriminatively across three dimensions: timesteps, the specific components of the embedding, and the depth of Cross-Attention layers, which are detailed in (b), (c), and (d), respectively. We perform perturbation across each of the three dimensions and evaluate rewards with BoN. Specifically: (b) illustrates that perturbation at early timesteps yields minimal improvement, while perturbation in later timesteps provides a significant boost. (c) indicates that the unconditional text embedding exhibits greater tolerance for perturbation compared to the conditional text embedding. (d) shows that shallower layers demonstrate higher tolerance for perturbation than deeper layers. \textbf{We use SD3.5 with ImageReward for evaluations, and more backbones' results are in Appendix.}}
    \vspace{-10pt}
    \label{fig:only_tep}
\end{figure}

\subsection{Frequency-Domain Analysis of Randomness}
\label{sec:frequency}
In TTS methods, the randomness stems from three sources: initial noise, SDE-injected noise, and resampling noise.  We call them \textbf{spatial noise randomness}, which acts in latent space and adheres to noise schedules. SDE-injected noise is taken as an example to analyze its impact on generation.
\vspace{-6pt}
\paragraph{Spatial Noise Randomness Itself.} We switch the sampling process from SDE to ODE at specific steps and employ the BoN method to select the highest-quality output to observe the impact of spatial noise randomness in TTS methods. As shown in Fig. \ref{fig:sde2ode}, in early denoising steps, this format of randomness effectively improves the upper bound of generation quality, but in later steps, it significantly degrades the generative quality.
We further investigate the influence of different frequency components in SDE-injected noise on generation quality. The results in Fig. \ref{fig:frequency} show that low-frequency components consistently have a positive effect, while partial removal of high-frequency components can enhance generation quality. This reveals the shifting role of spatial noise randomness throughout the gneration process from a frequency-domain perspective.

\begin{figure}[t]
    \centering
    \subfloat[AestheticScore distribution.\label{fig:distribution}]{
        \includegraphics[width=0.49\linewidth]{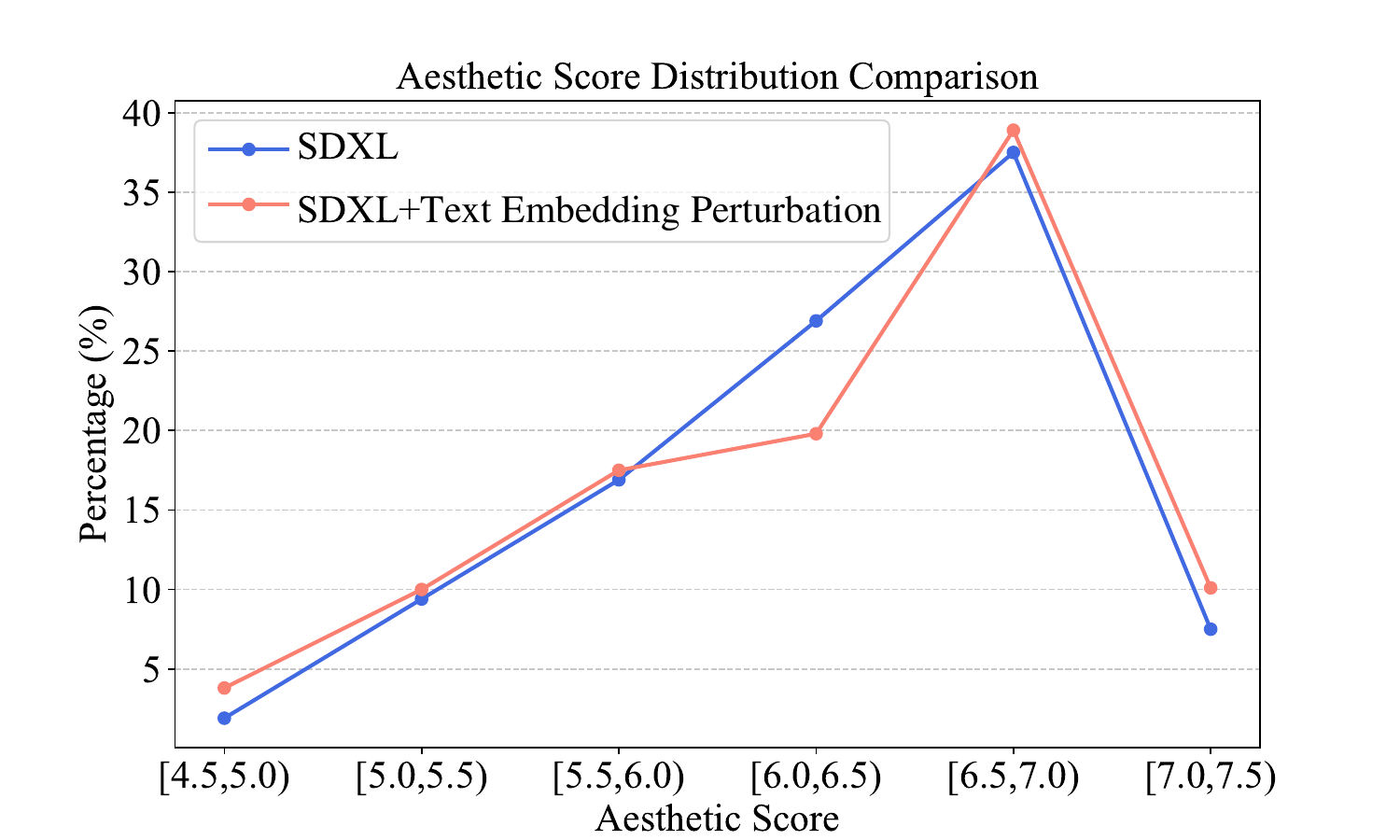}
    }
    \hfill
    \subfloat[Visual quality comparison.\label{fig:visual_quality}]{
        \includegraphics[width=0.46\linewidth]{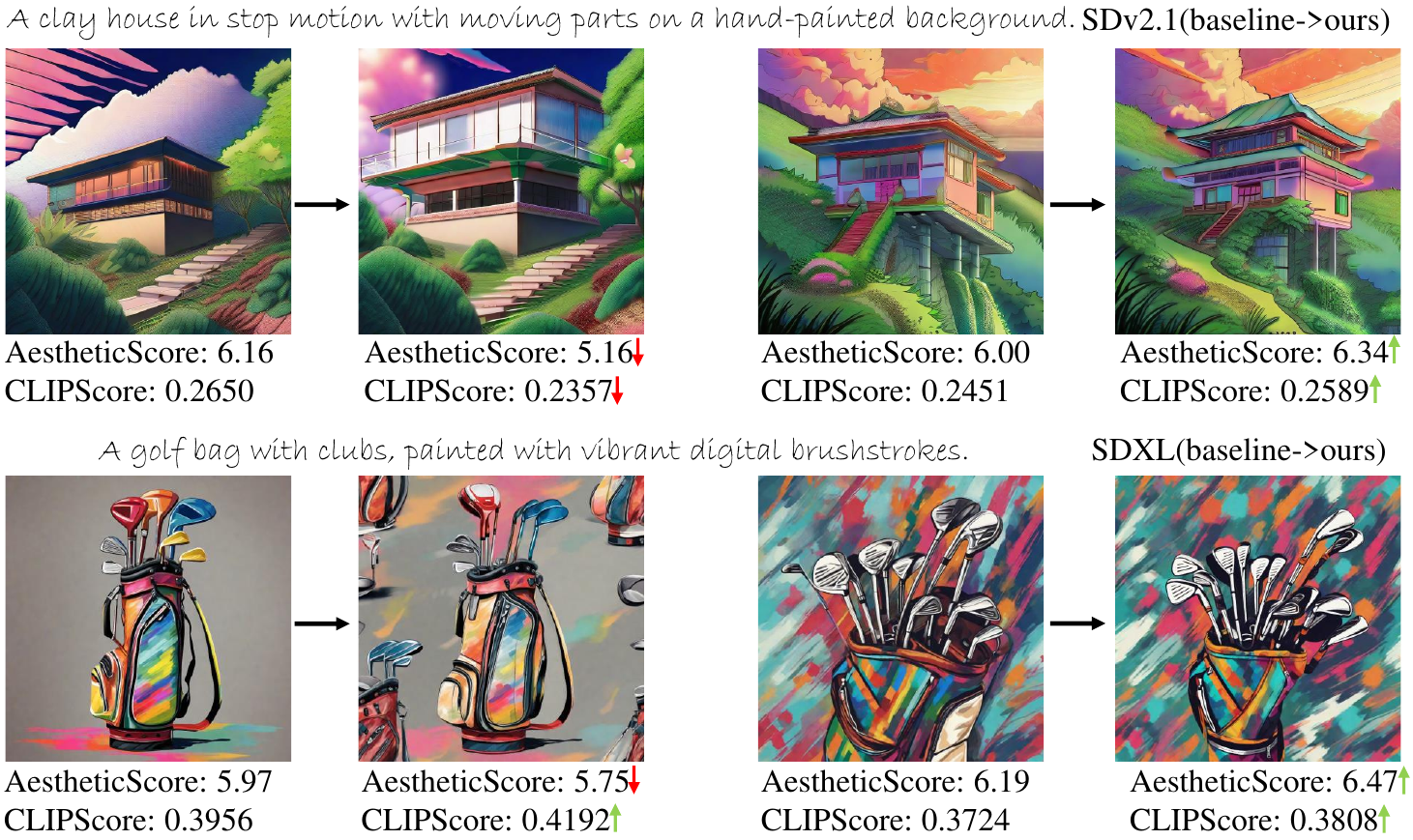}
    }
    \vspace{-1pt}
    \caption{\textbf{Text embedding perturbation effectively guarantees visual quality and text alignment in generated images.} As shown in (a), introducing controlled randomness through discriminative perturbation expands the performance range, yielding higher aesthetic peaks while reducing subpar outputs, thus enabling more efficient high-reward potential selection. Examples in (b) further demonstrate its capability to enhance visual quality (AestheticScore) and text alignment (CLIPScore) simultaneously. \textbf{Here we use SD3.5 to generate examples and conduct analysis.}}
    \vspace{-10pt}
\end{figure}

\paragraph{Spatial Noise Randomness and Text Embedding Perturbation.} We specifically compare the impact of these two format of randomness on generation, where we introduce only one type of randomness at the current step while keeping all other steps deterministic. As shown in Fig. \ref{fig:influence}, we measure the difference between introducing the randomness and not introducing it (via MSE) and find that spatial noise randomness primarily affects low-frequency components in early steps, while text embedding perturbation has a stronger influence on high-frequency details in later steps (also illustrated in Fig.~\ref{fig:tep2ntep}).
Fig. \ref{fig:diversity_visual} further shows that spatial noise randomness significantly shapes low-frequency structural elements (e.g., composition) in early stages but has minimal effect on fine details in later steps. In contrast, text embedding perturbation induces substantial variations in high-frequency details, leading to richer diversity in fine-grained features.
This also explains why CADS is not suitable for TTS methods: the excessive perturbation introduced in the early stages does not align with the preferred behavior of text embedding perturbation.
\begin{figure*}[t]
    \centering
    \subfloat[Text Embedding Perturbation Framework.\label{fig:framework}]{
        \includegraphics[width=0.9\textwidth]{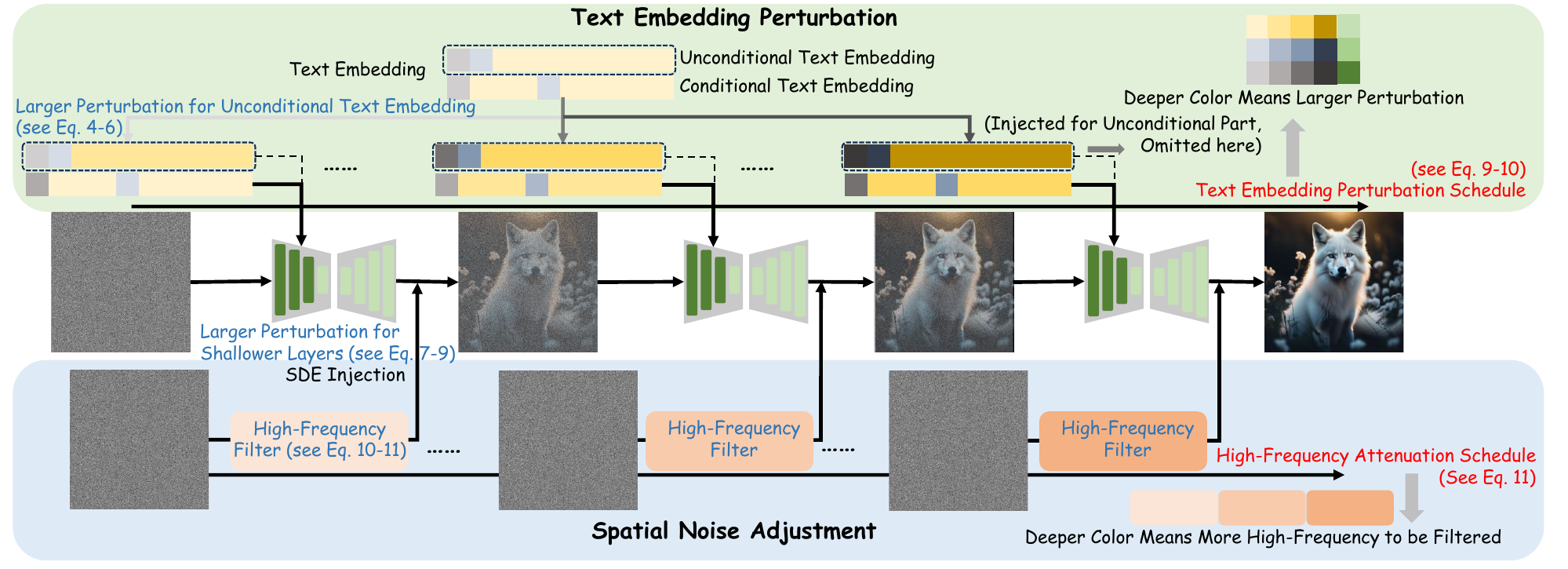}
    }
    \\ 
    \subfloat[When and where to adopt Text Embedding Perturbation.\label{fig:position}]{
        \includegraphics[width=0.9\textwidth]{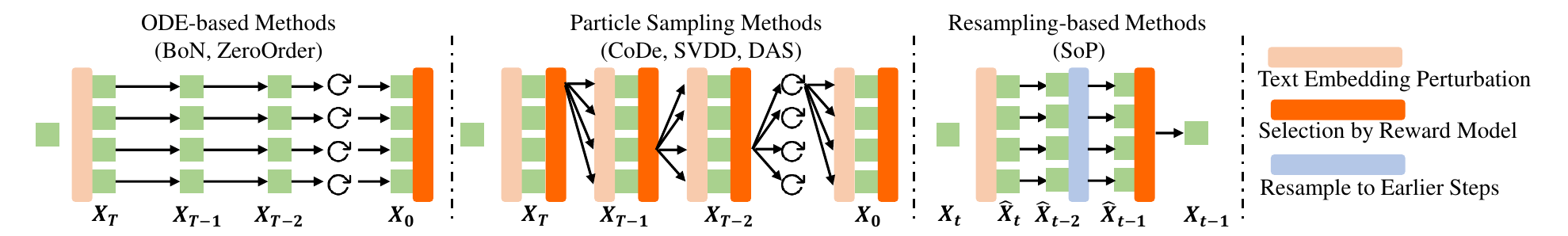}
    }
    \vspace{-5pt}
    \caption{\textbf{Our proposed TEP framework.} (a) illustrates the two main designs. Spatially, we apply specific perturbations to text embeddings according to their tolerance of perturbations, with stronger perturbations introduced to the unconditional text embedding and text embeddings in shallower layers. Meanwhile, we filter high-frequency components from the SDE-injected noise and restore it to standard Gaussian noise. Temporally, we progressively intensify the text embedding perturbation intensity throughout the denoising process, while simultaneously filtering more high-frequency content from the SDE-injected noise to enhance their complementarity. (b) demonstrates the application of our TEP framework across three different categories of TTS methods. In general, these methods introduce randomness or filter samples at certain timesteps, and we additionally inject text embedding perturbation at the same locations.}
    \vspace{-5pt}
    \label{fig:both}
\end{figure*}
It is worth noting that these two types of randomness exhibit strong complementarity throughout the denoising process: spatial noise randomness mainly influences the early stages of generation, while text embedding perturbation primarily affects the later stages. As a result, they can be effectively combined to further enhance both the variability (Fig. \ref{fig:influence}) and diversity (Fig. \ref{fig:diversity_visual}) of the generated outputs.

\subsection{Discriminative Perturbation of Text Embeddings in Diffusion Models}
As text embedding perturbation effectively enhances generative diversity and complements spatial noise randomness, the critical focus of our work is determining the specific format of perturbation required to maintain generative quality. Through experimentation, we discover the differential characteristics of text embedding perturbation across three dimensions, as shown in Fig. \ref{fig:differentiation}.

\paragraph{Better Performance in the Later Stages of Generation.} As illustrated in Fig. \ref{fig:tep2ntep}, we continually introduce perturbation up to a specific step and employ BoN. Text embedding perturbation shows minimal improvement in rewards during the initial stages, but yields a significant increase in the reward during the later generation stages. This observation suggests that this type of perturbation is more effective when applied with higher intensity in the latter steps.
\vspace{-5pt}
\paragraph{Greater Perturbation Tolerance of Unconditional Text Embeddings.} As depicted in Fig. \ref{fig:uncond_cond}, a substantial difference exists in the perturbation tolerance between the conditional and unconditional text embeddings. We independently perturb both and track the corresponding reward trend. As perturbation intensity steadily increases, the reward for the conditional text embedding exhibits an early decline, whereas the unconditional text embedding's reward only begins to decrease after the perturbation reaches a very large magnitude. This finding aligns with intuition: the conditional text embedding represents the target generation region, while the unconditional one defines the generation region to be avoided, and the target generation region is inherently much smaller than the avoidance generation region.

\paragraph{Greater Perturbation Tolerance of Shallower Layers in the Same Step.} As shown in Fig. $\ref{fig:layer_perturb}$, we divide the denoising network into two halves—deep layers and shallow layers—and independently perturb them to observe the reward trend. Perturbing the text embeddings within the cross-attention modules of the deep layers rarely yields a significant increase in the reward, while perturbation in the shallow layers quickly results in a rise in the reward. This indicates that the deep features are utilized for semantic restoration; consequently, corrupting the semantics leads to a degradation in denoising performance, a phenomenon previously demonstrated in numerous studies \cite{loos2025latent,wang2025dynamic,park2024explaining}.
\input{tab/sdxl}

\paragraph{Unified Discriminative Perturbation.} Combining the above three points of analysis, we simultaneously perform discriminative perturbation across the three dimensions (larger perturbation to shallower layers, later steps, and unconditional text embedding). We generate multiple images and present the range distribution of their rewards and corresponding visualizations. As shown in Fig. \ref{fig:distribution}, this form of perturbation maintains image generation quality effectively and leads to a higher reward ceiling, which is a highly desirable characteristic for TTS tasks. Furthermore, some images exhibit better detail and visual quality after text embedding perturbation, as illustrated in Fig. \ref{fig:visual_quality}. Consequently, we have successfully identified an appropriate text embedding perturbation strategy.

%% file: tab/sdxl.tex
\begin{table*}[h]
\centering
\caption{\textbf{Results of test-time scaling methods w/ and w/o our framework on SDXL.}}
\scalebox{0.89}{

\begin{tabular}{ccccccc}
\toprule
Sampling Method &w/ TEP & HPSv2$\uparrow$ & ImageReward$\uparrow$ & CLIPScore$\uparrow$ &AestheticScore$\uparrow$ &GenEval$\uparrow$\\ \hline\hline
None(SDXL)& \ding{55} &0.269&0.221&1.012&5.823&0.53\\ \hline\hline
\multicolumn{7}{c}{ODE-Based Methods} \\ \hline\hline
\multirow{2}{*}{BoN~\cite{ma2025inference}} &\ding{55} &0.284&0.943&1.034&6.392&0.59\\
&\ding{51} &\textbf{0.294}&\textbf{1.105}&\textbf{1.047}&\textbf{6.525}&\textbf{0.63}\\ \hline
\multirow{2}{*}{ZeroOrder~\cite{ma2025inference}} &\ding{55} &0.283&0.939&1.030&6.321&0.55\\
&\ding{51} &\textbf{0.293}&\textbf{0.977}&\textbf{1.044}&\textbf{6.448}&\textbf{0.62}\\ \hline\hline
\multicolumn{7}{c}{Particle Sampling Methods} \\ \hline\hline
\multirow{2}{*}{CoDe~\cite{singh2025code}} &\ding{55} &0.282&0.987&1.033&6.414&0.64\\
&\ding{51} &\textbf{0.301}&\textbf{1.328}&\textbf{1.068}&\textbf{6.704}&\textbf{0.76}\\ \hline
\multirow{2}{*}{SVDD~\cite{li2024derivative}} &\ding{55} &0.284&0.974&1.032&6.398&0.65\\
&\ding{51} &\textbf{0.299}&\textbf{1.303}&\textbf{1.063}&\textbf{6.722}&\textbf{0.78}\\ \hline
\multirow{2}{*}{DAS~\cite{kim2025test}} &\ding{55} &0.285&1.002&1.045&6.553&0.65\\
&\ding{51} &\textbf{0.302}&\textbf{1.379}&\textbf{1.084}&\textbf{6.836}&\textbf{0.75}\\ \hline\hline
\multicolumn{7}{c}{Resampling-Based Methods} \\ \hline\hline
\multirow{2}{*}{SoP~\cite{ma2025inference}} &\ding{55} &0.280&0.948&1.037&6.402&0.61\\
&\ding{51} &\textbf{0.288}&\textbf{1.032}&\textbf{1.070}&\textbf{6.536}&\textbf{0.66}\\ \hline

\bottomrule
\end{tabular}}

\label{tab:sdxl}

\end{table*}

%% file: text/4_method.tex
\section{Text Embedding Perturbation Framework}
\label{sec:method}
In this section, we fully present our Text Embedding Perturbation (TEP) framework. We elaborate in detail on our two design points in Fig. \ref{fig:framework} and explain the position and timing of incorporating text embedding perturbation in Fig.\ref{fig:position}.

\paragraph{Spatial Discriminative Adjustment for Both Randomness (Blue Color in Fig. \ref{fig:framework}).} For randomness brought by text embedding perturbation, we implement spatial discriminative perturbation by \textcircled{1} applying smaller perturbations to conditional text embeddings and larger perturbations to unconditional text embeddings:
\begin{equation}
\label{eq:cfg_perturb}
    \hat{\epsilon}_t = \epsilon_\theta(x_t, t, \hat{E}(y_\emptyset)) + w\left( \epsilon_\theta(x_t, t, \hat{E}(y_c)) - \epsilon_\theta(x_t, t, \hat{E}(y_\emptyset)) \right)
\end{equation}
\begin{equation}
    \hat{E}(y_\emptyset) = E(y_\emptyset)+w_1\epsilon_1
\end{equation}
\begin{equation}
    \hat{E}(y_c) = E(y_c)+w_2\epsilon_2
\end{equation}

Where $\epsilon_1, \epsilon_2 \sim N(0,I)$ and $w_1 >> w_2$. In this way, the model's output retains sufficient distinctiveness for high-reward potential selection even in the later denoising stages, while avoiding excessive disruption to semantics and guidance that would degrade visual quality and text faithfulness. Additionally, we implement finer-grained perturbation for conditional text embeddings: semantic embeddings (tokens before [EOS]) receive minimal perturbation to preserve semantic integrity, while padding embeddings (tokens after [EOS]) undergo stronger perturbation (still weaker than $w_1$) to ensure sufficient diversity. 
\textcircled{2} applying smaller perturbations to deeper layers and larger perturbations to shallower layers. Specifically, the interaction between the noisy latent and the text embedding takes place in the Cross Attention:
\begin{equation}
    \text{output} = \operatorname{CrossAttn}_i(x_t,E(y))
\end{equation}
Where, $\operatorname{CrossAttn}_i$ represents the Cross Attention of the $i$-th block. For stability, we simply set a threshold $k$ and consider layers before the $k$-th layer as shallow layers, and the remaining as deep layers. We then assign different relative coefficients for the perturbation applied to different layers:
\begin{equation}
    \text{output} = \operatorname{CrossAttn}_i(x_t,E(y)+s_iw\epsilon)
\end{equation}
\begin{equation}
    s_i = 
\begin{cases} 
1.5 & \text{if } i < k \\ 
0.5 & \text{if } i \geq k 
\end{cases}
\end{equation}
Where, for both the conditional text embedding and the unconditional text embedding, we use the same $s_i$.
For spatial noise randomness, we attenuate certain high-frequency components. Specifically, we first convert the signal to the frequency domain via Fast Fourier Transform (FFT). We then apply low-pass filtering with threshold $p$. Finally, we convert it back to the spatial domain:
\begin{equation}
    \epsilon_{low} = \operatorname{IFFT}(\operatorname{F_{low}}(\operatorname{FFT}(\epsilon),p))
\end{equation}
Where, $F$ denotes a low-pass filter with a constant threshold $p$.
To ensure it follows a standard Gaussian distribution, we renormalize it to obtain the final injected noise:
\begin{equation}
    \epsilon_{SDE} = (\epsilon_{low}-\operatorname{mean}(\epsilon_{low}))/\operatorname{std}(\epsilon_{low})
\end{equation}
\input{tab/flow}
\paragraph{Temporal Scheduling for Better Complementarity of Both Randomness (Red Color in Fig. \ref{fig:framework}).} First, considering that text embedding perturbation has a greater impact on high-frequency details in later stages and contributes more to high-frequency refinement, we progressively intensify this perturbation throughout the denoising process. Specifically, we parameterize perturbation weights $w_1$ and $w_2$ as monotonically increasing functions of $t$:
\begin{equation}
    \hat{E}(y_\emptyset) = E(y_\emptyset)+s_iw_1(t)\epsilon_1
\end{equation}
\begin{equation}
    \hat{E}(y_c) = E(y_c)+s_iw_2(t)\epsilon_2
\end{equation}
For spatial noise randomness, considering its negative impact on generation quality in later stages (primarily caused by its high-frequency components), we progressively increase the attenuation of these high-frequency elements throughout the denoising process:
\begin{equation}
    \epsilon_{low} = \operatorname{IFFT}(\operatorname{F_{low}}(\operatorname{FFT}(\epsilon),p(t)))
\end{equation}

\paragraph{The Position and Timing of Incorporating Text Embedding Perturbation.} Essentially, in TTS, randomness is introduced for sampling and selection. Simply put, for existing sampling-based TTS methods, we add perturbations to the text embedding before their sampling steps. Specifically, as shown in Fig. \ref{fig:position}, we categorize these methods and discuss the integration accordingly.
For ODE-based methods, which sample the initial noise only at the start and filter after denoising is complete, we perturb the text embedding only during the initial stage.
 For particle sampling methods, which rely on the randomness introduced by the SDE process, we perturb the text embedding again before each SDE sampling step. For resampling-based methods, we perturb the text embedding again at each resampling step.

%% file: tab/flow.tex
\begin{table*}[t]
\centering
\caption{\textbf{Results of test-time scaling methods w/ and w/o our framework on flow models(SD3.5).}}
\scalebox{0.9}{

\begin{tabular}{ccccccc}
\toprule
Sampling Method &w/ TEP & HPSv2$\uparrow$ & ImageReward$\uparrow$ & CLIPScore$\uparrow$ & AestheticScore$\uparrow$ & GenEval$\uparrow$ \\ \hline\hline
None(SD3.5) & \ding{55} &0.283 &0.547 &1.031 &5.602 &0.63 \\ \hline\hline
\multicolumn{6}{c}{ODE-Based Methods} \\ \hline\hline
\multirow{2}{*}{BoN~\cite{ma2025inference}} &\ding{55} &0.292 &0.992 &1.035 &6.271 &0.69\\
&\ding{51} &\textbf{0.301} &\textbf{1.102} &\textbf{1.041} &\textbf{6.357} &\textbf{0.71}\\ \hline
\multirow{2}{*}{ZeroOrder~\cite{ma2025inference}} &\ding{55} &0.299 &0.941 &1.078 &6.288 &0.67\\
&\ding{51} &\textbf{0.302} &\textbf{1.108} &\textbf{1.095} &\textbf{6.520} &\textbf{0.71}\\ \hline\hline
\multicolumn{7}{c}{Particle Sampling Methods} \\ \hline\hline
\multirow{2}{*}{CoDe~\cite{singh2025code}} &\ding{55} &0.296 &1.038 &1.077 &6.319 &0.75\\
&\ding{51} &\textbf{0.310} &\textbf{1.411} &\textbf{1.106} &\textbf{6.646} &\textbf{0.80}\\ \hline
\multirow{2}{*}{SVDD~\cite{li2024derivative}} &\ding{55} &0.301 &1.358 &1.106 &6.608 &0.69\\
&\ding{51} &\textbf{0.316} &\textbf{1.582} &\textbf{1.138} &\textbf{6.982} &\textbf{0.75}\\ \hline
\multirow{2}{*}{DAS~\cite{kim2025test}} &\ding{55} &0.294 &0.907 &1.107 &6.281 &0.76\\
&\ding{51} &\textbf{0.307} &\textbf{1.270} &\textbf{1.116} &\textbf{6.403} &\textbf{0.81}\\ \hline
\multicolumn{6}{c}{Resampling-Based Methods} \\ \hline\hline
\multirow{2}{*}{SoP~\cite{ma2025inference}} &\ding{55} &0.293 &0.984 &1.040 &6.010 &0.69\\
&\ding{51} &\textbf{0.301} &\textbf{1.077} &\textbf{1.056} &\textbf{6.185} &\textbf{0.72}\\ \hline

\bottomrule
\end{tabular}}

\label{tab:flow}
\end{table*}

%% file: text/5_exp.tex
\section{Experiment}
\label{sec:exp}
Here, we demonstrate the applications of our TEP framework on T2I generation. 
More extended \textbf{applications} (i.e., T2V generation) and \textbf{computation} are shown in \textbf{Appendix}.

\begin{figure}[t]
	\centering
	\includegraphics[width=1\linewidth]{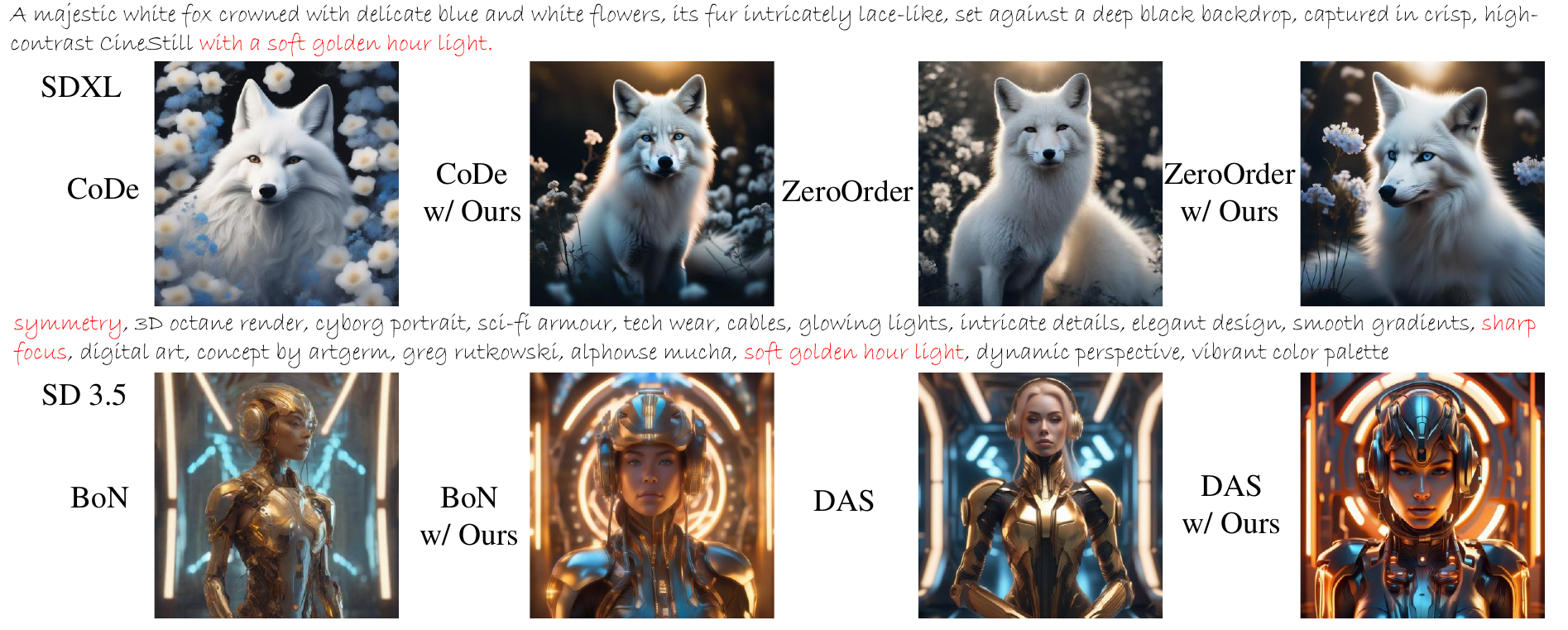}
	\caption{Visual results of baselines w/ and w/o our methods.} 
	\label{fig:results}
\end{figure}

\subsection{Experimental settings}

\noindent\textbf{Baselines.} We integrate TEP with existing TTS methods, including ODE-based methods (BoN, ZeroOrder~\cite{ma2025inference}), particle sampling methods (SVDD~\cite{li2024derivative}, CoDe~\cite{singh2025code}, DAS~\cite{kim2025test}), and resampling-based methods (SoP~\cite{ma2025inference}).

\noindent\textbf{Backbones and their Text Encoders.} We evaluate a diverse range of T2I diffusion models with different text encoders: \textbf{SD2.1}~\cite{Rombach_2022_CVPR} (CLIP), \textbf{SDXL}~\cite{podell2023sdxl} (dualCLIP), \textbf{SD3.5}~\cite{esser2024scaling} (dualCLIP+T5), \textbf{Flux}~\cite{flux2024} (dualCLIP+T5), and \textbf{show-o2}~\cite{xie2025show} (VLM). We present the results for the most widely adopted models, \textbf{SDXL} and \textbf{SD3.5}, in the main text, with others' results fully documented in \textbf{Appendix}.

\noindent\textbf{Evaluations.} We conduct experiments on \textbf{Open-Image-Pref-v1 datasets with 7k+ prompts} and \textbf{GenEval} benchmarks. We ensure that the number of simultaneously denoised latents at each step is fixed at 16.

\noindent\textbf{Reward Models.}  We test our method on ImageReward~\cite{xu2023imagereward}, HPSv2~\cite{wu2023human}, CLIPScore~\cite{hessel2021clipscore}, and AestheticScore~\cite{Schuhmann:aesthetics}. These reward models serve both as intermediate verifiers and held-out rewards (final rewards). We additionally apply GenEval~\cite{ghosh2023geneval} as the held-out reward for ImageReward verifier. More information is shown in \textbf{Appendix}.

\input{tab/ablation_1}

\subsection{Main results}
\paragraph{Results on Unet-Based T2I Diffusion Models.} In this task, we use the same reward model as verifiers and held-out rewards. Results on GenEval are added for object-focused evaluation. As shown in Tab. \ref{tab:sdxl}, our method demonstrates significant improvements when seamlessly integrated with existing TTS methods. Among these, the enhancement is most pronounced for particle sampling methods, given that they not only conduct thorough searches of sampling paths but also extensively explore noise directions for generation.
\vspace{-9pt}
\paragraph{Results on Flow-Based T2I Diffusion Models.} For flow models, we replace the default ODE solver with an SDE process following~\citet{liu2025flow}, then apply these TTS methods along with our framework. As shown in Tab. \ref{tab:flow}, in flow-based models, our framework also demonstrates strong performance, achieving improvements across all baselines. We present complete tables and additional results on flow-based diffusion models in \textbf{Appendix}.

\subsection{Ablation study}

\paragraph{Ablations on Components and their Computation.} Our framework involves processing both text embedding perturbation and spatial noise randomness, so we conduct ablation studies on these components. We perform experiments using CoDe on SD 2.1 and evaluate with ImageReward and HPSv2. As shown in Tab. \ref{tab:ablation}, each component contributes to improved generation quality and only brings negligible additional computation.
More ablations on framework design and hyperparameter settings are presented in \textbf{Appendix}.

\vspace{-6pt}

\input{tab/generalization}

\vspace{-3pt}

\begin{figure}[t]
    \centering
    \subfloat[Scalability of NDFEs.\label{fig:ndfe}]{
        \includegraphics[width=0.46\linewidth]{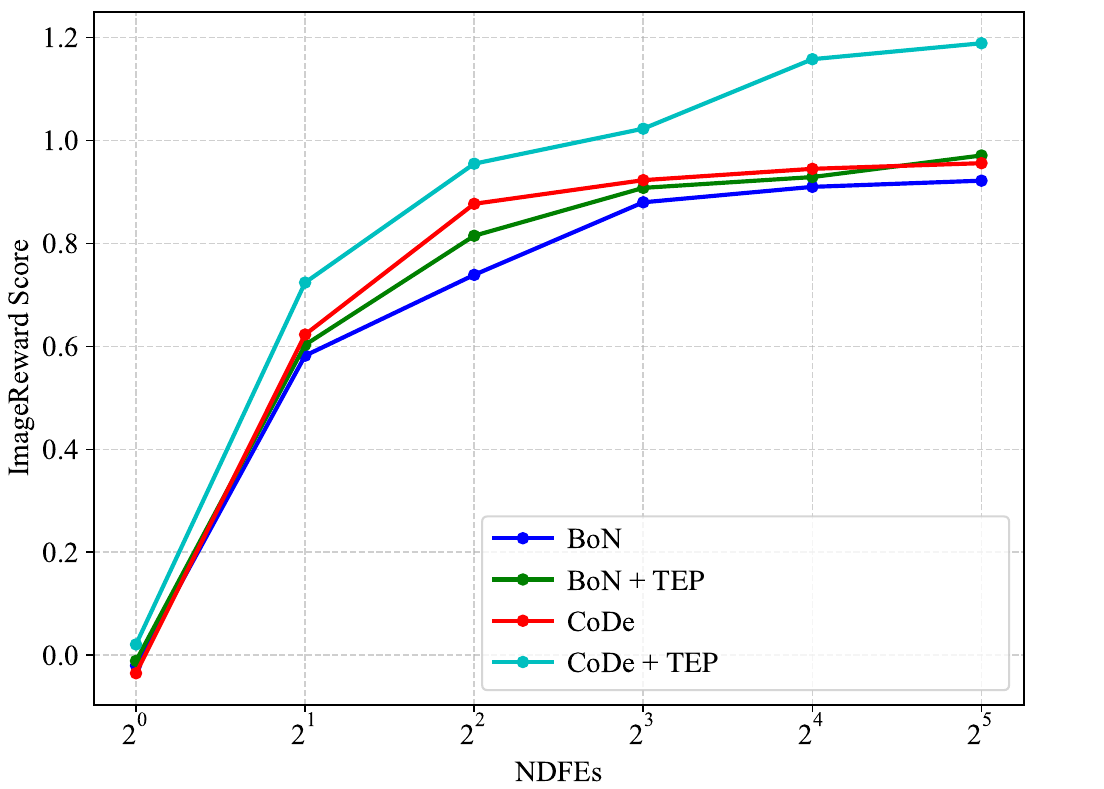}
    }
    \hfill
    \subfloat[Scalability of NRFEs.\label{fig:nrfe}]{
        \includegraphics[width=0.46\linewidth]{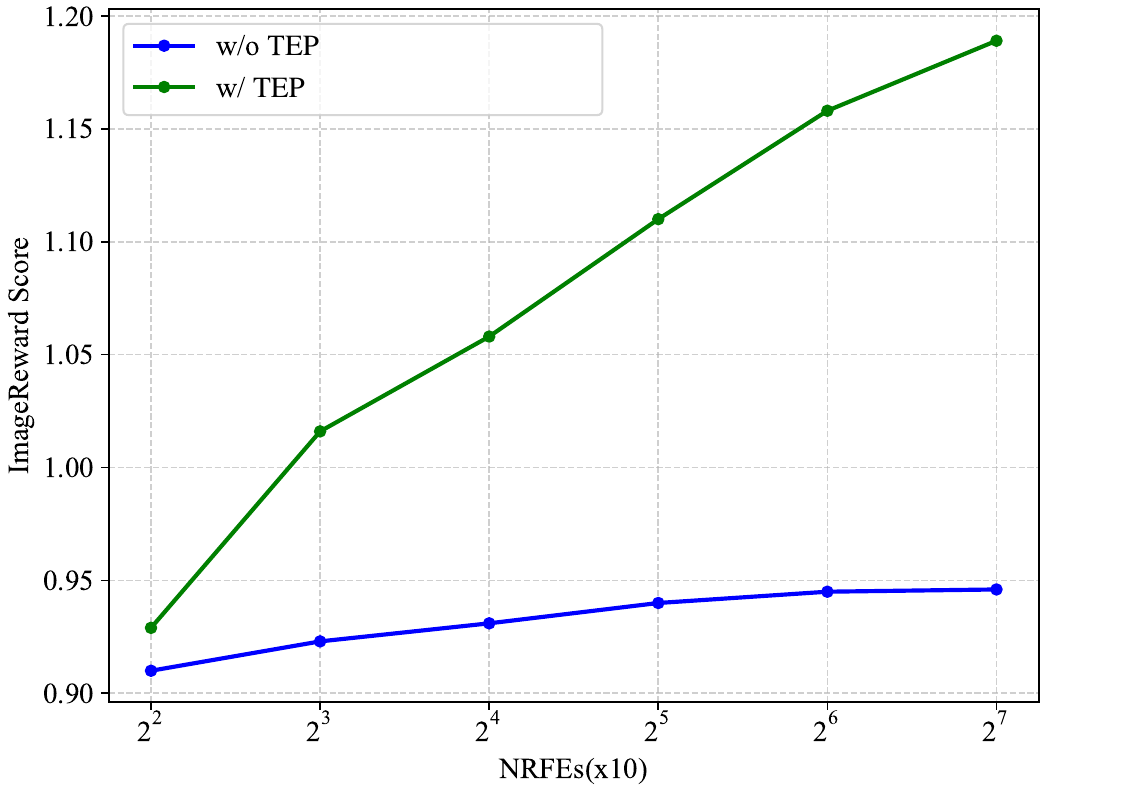}
    }
    \vspace{-6pt}
    \caption{The change curve of ImageReward with increasing NDFEs and NRFEs. In (a), our method helps the model achieve a higher upper limit and maintains an upward trend even with more NDFEs. Notably, although CoDe and BoN have same NDFEs, CoDe introduces additional randomness and evaluations, resulting in better performance. So we use NRFE in (b) to show the benefits of more evaluations during generation, where the advantage of our framework becomes increasingly evident as NRFE grows.}
    \vspace{-6pt}
\end{figure}

\paragraph{Num of NDFEs and NRMEs for Generation.} In Fig. \ref{fig:ndfe}, we demonstrate how NDFE (number of denoising function evaluations) and NRFE (number of reward function evaluations) influence the generation process. NDFE is related to the number of sampling steps, the initial noise quantity, and the number of intermediate sampling particles. Among these, increasing the number of sampling steps has a minor impact on the generation metrics, whereas raising the initial noise quantity or the number of particles can help the generation metrics rise rapidly until they converge to a stable level. However, particle sampling methods often outperform ODE-based methods, even when their NDFE values are the same. Therefore, we introduce NRFE, which also represents the number of filtering steps the model performs during TTS. In Fig. \ref{fig:nrfe}, we observe that as NRFE increases, the performance of the baseline gradually improves and then converges to a stable level, indicating that the model transitions from ODE-based methods to particle sampling methods. Interestingly, after incorporating text embedding perturbation, the generation metrics of the model continue to rise, reaching a higher upper limit. This is easy to understand: NRFE also represents the number of perturbations we introduce, and these perturbations provide models with a larger search space, accompanied by a higher upper bound. As NRFE increases, the model's search becomes more thorough, allowing it to better approach this upper limit.
\vspace{-6pt}
\paragraph{Generalizations on Text Encoders.} The strong generalizability of our approach lies in the fact that applying a reasonable perturbation to the text embedding derived from any text encoder consistently leads to performance improvements for diffusion models in TTS tasks, as shown in Tab. \ref{tab:generalization}. We utilize CoDe methods w/ and w/o our TEP, and use ImageReward as our evaluation metric.

\vspace{-3pt}

%% file: tab/ablation_1.tex
\begin{table}[t]
    \centering
\caption{\textbf{Ablations of TEP.}}
\vspace{-6pt}
\scalebox{0.9}{

        \begin{tabular}{cccc}
            \toprule
            Perturbation & IR$\uparrow$ & HPSv2$\uparrow$ &Time\\
            \hline
            None &0.987  &0.282  &15.488s\\
            +Text Embedding Perturbation &1.253  &0.297 &15.645s\\
            +Spatial Noise Adjustment &1.148 &0.288 &15.703s\\
            Ours &\textbf{1.328} &\textbf{0.301} &15.732s\\
            \bottomrule
        \end{tabular}}

        \label{tab:ablation}

\end{table}

%% file: tab/generalization.tex
\begin{table}[t]
    \centering
\caption{\textbf{Generalization of different text encoders.}}
\vspace{-6pt}
\scalebox{0.9}{

        \begin{tabular}{cccc}
            \toprule
            Model &Text Encoder & w/o TEP & w/ TEP\\
            \hline
            SD2.1 &CLIP &0.945  &\textbf{1.158}  \\
            SDXL &dualCLIP &0.987  &\textbf{1.328} \\
            SD3.5 &dualCLIP+T5 &1.038 &\textbf{1.411}  \\
            Show-O2 &VLM &1.215 &\textbf{1.448} \\
            \bottomrule
        \end{tabular}}

        \label{tab:generalization}
    \vspace{-10pt}
\end{table}

%% file: text/6_conclusion.tex
\section{Conclusion}
\label{sec:conclusion}
We introduce a novel format of randomness for TTS in T2I diffusion models, text embedding perturbation, which helps improve generative quality, significantly enhancing both the visual quality and textual fidelity of generated images. While T2I diffusion models continue to advance with remarkable generative potential, current TTS methods may struggle to fully exploit these capabilities. We highlight the importance of further exploration of TTS for T2I diffusion models to fully unlock their potential.